\documentclass[letterpaper, 10 pt, conference]{ieeeconf}  

\IEEEoverridecommandlockouts                              

\usepackage{graphicx}
\usepackage{multicol}
\usepackage{tikz}
\usepackage{comment}
\usepackage{amsmath,amssymb} 
\usepackage{color}
\usepackage{orcidlink}
\usepackage{multirow}
\usepackage{adjustbox}
\usepackage{threeparttable}
\usepackage{cuted}
\usepackage{capt-of}
\usepackage{dblfloatfix}

\title{\LARGE \bf
TransNet: Transparent Object Manipulation\\ Through Category-Level Pose Estimation
}

\author{
Huijie Zhang$^{1}$
\and
Anthony Opipari$^{1}$
\and
Xiaotong Chen$^{1}$
\and
Jiyue Zhu$^{1}$
\and
Zeren Yu$^{2}$
\and
Odest Chadwicke Jenkins$^{1}$
\thanks{$^{1}$ H. Zhang, A. Opipari, X. Chen, J. Zhu, and O. Jenkins are with the Electrical Engineering and Computer Science, Mechanical Engineering, and Robotics departments, University of Michigan, Ann Arbor, MI, USA.
}
\thanks{$^{2}$ Z. Yu is with Department of Computer Science, University of Southern California, Los Angeles, CA, USA.}
}
\begin{document}

\maketitle
\thispagestyle{empty}
\pagestyle{empty}


\begin{strip}
\vspace{-2.5cm}
\begin{center}
    \includegraphics[width=1.0\textwidth]{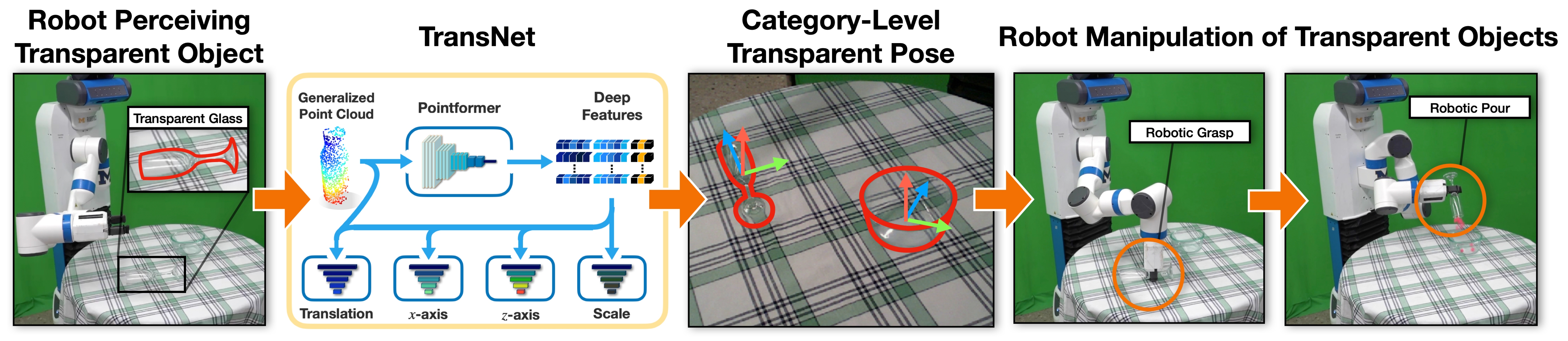}
    \captionof{figure}{TransNet is a pipeline for category-level transparent object pose estimation and manipulation. Given an RGB-D image of a scene containing transparent objects, TransNet estimates the 6D pose and 3D scale of each object in the scene for use in downstream manipulation tasks such as grasping and pouring.}
    \label{fig:teaser}
\end{center}
\vspace{-0.6cm}
\end{strip}

\begin{abstract}
Transparent objects present multiple distinct challenges to visual perception systems.
First, their lack of distinguishing visual features makes transparent objects harder to detect and localize than opaque objects. Even humans find certain transparent surfaces with little specular reflection or refraction, like glass doors, difficult to perceive. 
A second challenge is that depth sensors typically used for opaque object perception cannot obtain accurate depth measurements on transparent surfaces due to their unique reflective properties. 
Stemming from these challenges, we observe that transparent object instances within the same category, such as cups, look more similar to each other than to ordinary opaque objects of that same category. 
Given this observation, the present paper explores the possibility of category-level transparent object pose estimation rather than instance-level pose estimation. We propose \textit{\textbf{TransNet}}, a two-stage pipeline that estimates category-level transparent object pose using localized depth completion and surface normal estimation. TransNet is evaluated in terms of pose estimation accuracy on a large-scale transparent object dataset and compared to a state-of-the-art category-level pose estimation approach. Results from this comparison demonstrate that TransNet achieves improved pose estimation accuracy on transparent objects. Moreover, we use TransNet to build an autonomous transparent object manipulation system for robotic pick-and-place and pouring tasks. The project webpage is available at \href{https://progress.eecs.umich.edu/projects/transnet/}{https://progress.eecs.umich.edu/projects/transnet/}.

\end{abstract}

\vspace{-0.2cm}

\section{Introduction}
\label{sec:intro}
From glass doors and windows to kitchenware and all kinds of containers, objects with transparent materials are prevalent throughout daily life.
Thus, perceiving the pose (position and orientation) of transparent objects is a crucial capability for autonomous perception systems seeking to interact with their environment.
However, transparent objects present unique perception challenges both in the RGB and depth domains.

For RGB, the color appearance of transparent objects is highly dependent on the background, viewing angle, material, and lighting condition, due to light reflection and refraction effects.
For depth, common commercially available depth sensors record mostly invalid or inaccurate depth values within the transparent region.
Such visual challenges, especially missing detection in the depth domain, pose severe challenges for autonomous object manipulation and obstacle avoidance tasks.
This paper addresses these challenges by studying how category-level transparent object pose estimation may be achieved using end-to-end learning.

Recent works have shown promising results on grasping transparent objects by completing the missing depth values followed by the use of a geometry-based grasp engine~\cite{sajjan2020clear,ichnowski2021dex,fang2022transcg}, transfer learning from RGB-based grasping neural networks~\cite{weng2020multi}, light-field feature learning~\cite{zhou2019glassloc}, or domain-randomized depth noise simulation~\cite{dai2022domain}.
For more advanced manipulation tasks such as rigid body pick-and-place or liquid pouring, geometry-based estimations, such as symmetrical axes, edges~\cite{phillips2016seeing} or object poses~\cite{lysenkov2013recognition, xu20206dof, dai2022domain}, are required to model the manipulation trajectories.
Instance-level transparent object poses could be estimated from keypoints on stereo RGB images~\cite{liu2020keypose,liu2021stereobj}, a light-field camera~\cite{zhou2018plenoptic, zhou2020lit}, or directly from a single RGB-D image~\cite{xu20206dof} with support plane assumptions.
Recently emerged large-scale transparent object datasets~\cite{sajjan2020clear,xu2021seeing,liu2021stereobj,fang2022transcg,chen2022clearpose} pave the way for addressing the problem using deep learning.

In this paper, we set out to extend the frontier of 3D transparent object perception by building upon recent work. We introduce \textit{TransNet}, a category-level pose estimation pipeline for transparent objects that outperforms a state-of-the-art baseline. We further explore the effect of input modalities, feature embedding methods, and depth-normal consistency through the learning process to provide insights for future research. Moreover, we build an autonomous transparent object manipulation system using TransNet and demonstrate its efficacy in pick-place and pouring tasks.

\section{Related Work}
\label{sec:related_works}

\subsection{Transparent Object Visual Perception for Manipulation}

Transparent objects must be perceived before they can be acted on.
Lai~\textit{et al.}~\cite{lai2015transparent} and Khaing~\textit{et al.}~\cite{khaing2018transparent} developed CNN models to detect transparent objects from RGB images.
Xie~\textit{et al.}~\cite{xie2020segmenting} proposed a deep segmentation model that achieved state-of-the-art accuracy.
ClearGrasp~\cite{sajjan2020clear} employed depth completion for use with pose estimation on robotic grasping tasks, where they trained three DeepLabv3+~\cite{chen2018encoder} models to learn transparency mask, surface normal, and boundary, respectively.
Follow-on studies developed different approaches for depth completion, including implicit functions~\cite{zhu2021rgb}, NeRF reconstruction~\cite{ichnowski2021dex}, combined point cloud and depth features~\cite{xu2021seeing}, adversarial learning~\cite{tang2021depthgrasp}, multi-view geometry~\cite{chang2021ghostpose}, RGB image completion~\cite{fang2022transcg}, and sim2real transfer~\cite{dai2022domain}.
Weng~\textit{et al.}~\cite{weng2020multi} used transfer learning from the RGB to the depth sensor domain without completing raw depth.
For instance-level pose estimation, Xu~\textit{et al.}~\cite{xu20206dof} utilized segmentation, surface normal, and image coordinate UV-map as input to a network similar to~\cite{tian2020robust} that can estimate 6 DOF object pose.
Keypose~\cite{liu2020keypose} proposed to estimate 2D keypoints and regress object poses from stereo images using triangulation.
For other special sensors, Xu~\textit{et al.}~\cite{xu2015transcut} used light-field images to perform segmentation using a graph-cut-based approach.
Kalra \textit{et al.}~\cite{kalra2020deep} trained Mask R-CNN~\cite{he2017mask} using polarization images as input to outperform the baseline that was trained on only RGB images by a large margin.
Zhou \textit{et al.}~\cite{zhou2018plenoptic,zhou2019glassloc,zhou2020lit} employed light-field images to learn features for robotic grasping and object pose estimation.
Along with the proposed methods, massive datasets, across different sensors and both synthetic and real-world domains, have been collected and made public for various related tasks~\cite{xie2020segmenting,sajjan2020clear,liu2020keypose,zhou2020lit,kalra2020deep,liu2021stereobj,zhu2021rgb,xu2021seeing,fang2022transcg,chen2022clearpose}.
Compared with these previous works, and to the best of the authors' knowledge, we propose the first category-level pose (6D pose + 3D scale) estimation approach developed specifically for transparent objects.

\subsection{Opaque Object Category-level Pose Estimation}

Category-level object pose estimation aims to estimate unseen objects' 6D pose within seen categories and their scale or canonical shape.
Wang \textit{et al.}~\cite{wang2019normalized} introduced the Normalized Object Coordinate Space (NOCS) for dense 3D correspondence learning and solved for object pose and scale using the learned NOCS map and masked depth~\cite{umeyama1991least}.
More recent works explored different aspects to improve pose estimation accuracy.
A category-level shape prior is found to be beneficial for pose estimation accuracy in~\cite{tian2020shape} and further improved in~\cite{chen2021sgpa,chen2021fs, di2022gpv}.
DualPoseNet~\cite{lin2021dualposenet}, 6D-ViT~\cite{zou20216d}, ACR-Pose~\cite{fan2021acr}, and CPPF~\cite{you2022cppf} proposed to incorporate rotation-invariant embedding, Transformer networks, Generative Adversarial Networks, and deep point-pair-feature, respectively.
However, these techniques require high-quality depth input provided by opaque objects with Lambertian light reflectance.
One recent work by Dai~\textit{et al.}~\cite{dai2022domain} proposed a data generation system that simulates the noise on non-Lambertian surfaces of active stereo depth cameras and demonstrated its usage in category-level pose estimation and robotic grasping.

Compared with these techniques, TransNet takes advantage of recent advances in transparent depth completion approaches and directly learns category-level pose estimation from imperfect depth predictions, carefully considering multi-modal input, feature embedding architecture, and depth-normal consistency.
\section{TransNet}
\label{sec:method}

\begin{figure*}[ht]
    \centering
    \includegraphics[width=\textwidth]{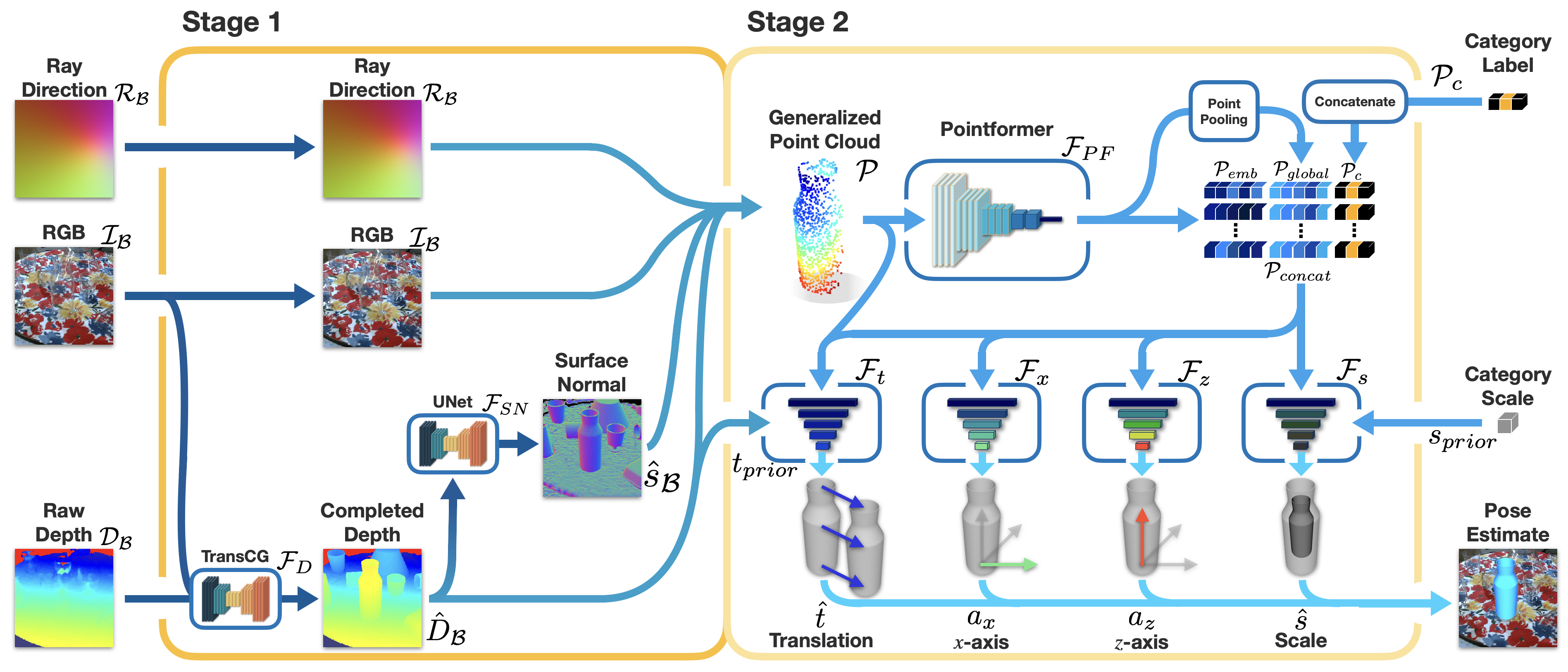}
    \caption{The two-stage architecture of TransNet. Image patches of RGB, ray direction, surface normal, and completed depth are concatenated and embedded to learn the translation, x-axis, z-axis, and scale separately of the estimated category-level pose.
    }
    \label{fig:arch}
    \vspace{-0.7cm}
\end{figure*}    

Given an input RGB-D pair ($\mathcal{I}$, $\mathcal{D}$) viewing a scene that includes transparent objects, our goal is to predict the objects' 6D rigid body transformations $[\textbf{R}|\textbf{t}]$, 3D scales $\textbf{s}$ in the camera coordinate frame and object classes within a predefined category set, where $\textbf{R} \in SO(3), \textbf{t} \in \mathbb{R}^{3}$ and $\textbf{s} \in \mathbb{R}^{3}_{+}$. 
In this problem, inaccurate or invalid depth readings exist within the image region of transparent objects (represented as a binary mask).
To solve the category-level pose estimation problem, we propose a novel two-stage deep neural network pipeline, \textit{TransNet}.

\subsection{Architecture Overview}

TransNet first applies a fine-tuned instance segmentation module to extract individual object patches from the overall image. As shown in Figure~\ref{fig:arch}, in the first stage, TransNet takes the patches as input and extracts multi-modal features corresponding to each transparent object instance. Depth completion and surface normal estimation are applied on RGB-D patches to obtain depth-normal pairs through cross-task consistency learning. The estimated depth-normal pairs, RGB, and ray direction patches are concatenated to feature patches, followed by a random sampling strategy within the instance masks to generate features referred to as the \textit{generalized point cloud}. In the second stage, the generalized point cloud is processed through Pointformer \cite{zou20216d} to produce embedded feature vectors. The pose is then separately estimated in four decoder modules for object translation, $x$-axis, $z$-axis, and scale, respectively.

\subsection{Object Instance Detection and Segmentation}

Similar to other category-level pose estimation work \cite{di2022gpv}, we fine-tune a Mask R-CNN \cite{he2017mask} model to obtain the object's bounding box $\mathcal{B}$, segmentation mask $\mathcal{M}$ and category label $\mathcal{P}_c$.
Patches of ray direction $\mathcal{R}_{\mathcal{B}}$, RGB $\mathcal{I}_{\mathcal{B}}$ and raw depth $\mathcal{D}_{\mathcal{B}}$ are extracted according to the bounding box $\mathcal{B}$ and serve as input to the first stage of TransNet. The UV mapping itself is an important cue when estimating poses from patches~\cite{jiang2022uni6d}, as it provides information about the relative position and size of the patches within the overall image.
We use ray direction instead of UV mapping because it also contains camera intrinsic information. Here the ray direction $\mathcal{R}$ represents the direction from camera origin to each pixel in the camera plane under the camera frame. For each pixel $(u,v)$:
\begin{align}
\begin{split}
    p &= \begin{bmatrix}u & v & 1\end{bmatrix}^T \\
    \mathcal{R} &= \frac{K^{-1} p}{\left\lVert K^{-1} p\right\rVert^2}
\end{split}
\end{align}
where $p$ is the homogeneous UV coordinate in the image plane and $K$ is the camera intrinsic matrix.

\subsection{Cross-task Consistency for Depth-Normal Pairs}



We apply a recent state-of-the-art approach, TransCG~\cite{fang2022transcg} for depth completion ($\mathcal{F}_{D}$) and U-Net~\cite{ronneberger2015u} for surface normal estimation ($\mathcal{F}_{SN}$).

\begin{align}
\begin{split}
\label{eq:dc&sne}
    \hat{\mathcal{D}}_\mathcal{B} &= \mathcal{F}_{D}\left(\mathcal{I}_\mathcal{B}, \mathcal{D}_\mathcal{B}\right)\\
    \hat{\mathcal{S}}_\mathcal{B} &= \mathcal{F}_{SN}\left(\hat{\mathcal{D}}_\mathcal{B}\right)
\end{split}
\end{align}
where the depth estimation $\hat{\mathcal{D}}_\mathcal{B}$ is used as input to the surface normal estimation module.
The two models are first trained separately with depth and surface normal losses, $\mathcal{L}_d$ and $\mathcal{L}_s$, and then trained together using cross-task consistency. Following \cite{zamir2020robust}, we design a consistency loss $\mathcal{L}_{con}$ to train both networks. Both $\mathcal{L}_d$ and $\mathcal{L}_s$ are implemented using $L_2$ loss for depth completion and surface normal estimation within transparent mask $\mathcal{M}$. 
\begin{align}
\begin{split}
\label{eq:loss}
    \mathcal{L}_d &= \frac{1}{N_p}\sum_{p\in\mathcal{M}}\left\lVert\hat{\mathcal{D}}_p -  \mathcal{D}^{*}_p\right\rVert^2 \\
    \mathcal{L}_s &= \frac{1}{N_p}\sum_{p\in\mathcal{M}}\left\lVert\hat{\mathcal{S}}_p -  \mathcal{S}^{*}_p\right\rVert^2 \\
    \mathcal{L}_{con} &= \frac{1}{N_p}\sum_{p\in\mathcal{M}}\left\lVert \hat{\mathcal{S}}_p - \mathcal{F}_{SN}(\mathcal{D}^{*}_\mathcal{B})_p \right\rVert^2
\end{split}
\end{align}
where $\mathcal{D}^{*}$ and $\mathcal{S}^{*}$ are the ground truth depth and surface normal images, and $N_p$ refers to the number of masked pixels, respectively. During training, these losses are used according to the perceptual loss training strategy described by Zamir et al.~\cite{zamir2020robust}.
Following cross-task consistency training, the depth completion and surface normal estimation networks are frozen and used to generate input for the second stage of TransNet.

\subsection{Generalized Point Cloud}
The generalized point cloud is implemented as a concatenation of multi-modal input features: $\mathcal{I}_\mathcal{B}, \;\mathcal{R}_\mathcal{B}, \;\hat{\mathcal{D}}_\mathcal{B},$ and $\hat{\mathcal{S}}_\mathcal{B}$. Specifically, we randomly sample $N$ pixels within the transparent mask $\mathcal{M}$ of the feature patch to obtain a generalized point cloud $\mathcal{P}\in \mathbb{R}^{N\times 10}$. The experiment described in Section \ref{exp:generalized point cloud} explores the best choice of features for use in the generalized point cloud.


\subsection{Transformer Feature embedding}

Given a generalized point cloud $\mathcal{P}$ corresponding to a detected object, TransNet employs Pointformer~\cite{zou20216d}, a multi-stage transformer-based point cloud embedding method to encode object features $\mathcal{P}_{emb} \in \mathbb{R}^{N\times d_{emb}}$.
\begin{equation}
\label{eq:emb}
    \mathcal{P}_{emb} = \mathcal{F}_{PF}\left(\mathcal{P}\right)
\end{equation}
Experiments described in Section~\ref{exp:embedding} evaluate the effectiveness of using alternative point cloud embedding methods, such as 3D-GCN~\cite{Lin_2020_3dgcn}, for use in the transparent object context with noisy depth data.

Next, a Point Pooling layer (a multilayer perceptron (MLP) with max-pooling) is applied to extract global features $\mathcal{P}_{global}$. The global features are then concatenated with local features $\mathcal{P}_{emb}$ and a one-hot category $\mathcal{P}_c$ label from the instance segmentation network. The concatenations result in a feature vector to be used as input to the decoder: 
\begin{align}
\begin{split}
\label{eq:concat}
    &\mathcal{P}_{global} = \text{MaxPool}\left(\text{MLP}\left(\mathcal{P}_{emb}\right)\right) \\
    &\mathcal{P}_{concat} = \left[\mathcal{P}_{emb}, \mathcal{P}_{global}, \mathcal{P}_c\right]
\end{split}
\end{align}

\subsection{Pose and Scale Estimation}

After the feature embeddings are extracted based on multi-modal input, four separate decoders are applied to estimate the object's translation, $x$-axis, $z$-axis, and scale. Similar to \cite{chen2021fs}.

\noindent\textbf{Translation Residual Estimation}
The translation decoder $\mathcal{F}_{t}$ learns a 3D translation residual from the object translation prior $t_{prior}$ calculated as the average of predicted 3D coordinate over the sampled pixels in $\mathcal{P}$. 
\begin{align}
\begin{split}
\label{eq:trans}
    t_{prior} &= \frac{1}{N_p}\sum_{p\in N} K ^{-1} \left[u_p \  v_p \  1\right]^T \hat{\mathcal{D}_p} \\
    \hat{t} &= t_{prior} + \mathcal{F}_{t}\left(\left[\mathcal{P}_{concat}, \mathcal{P}\right]\right) \\
\end{split}
\end{align}
where $u_p$, $v_p$ are the 2D pixel coordinate for the selected pixel. We use the $L_1$ loss between the ground truth and estimated position:
\begin{equation}
\label{eq:trans_loss}
    \mathcal{L}_t = \left\lvert\hat{t} - t^*\right\rvert
\end{equation}

\noindent\textbf{Rotation Estimation} 
We decouple the 3D rotation matrix $R$ into two orthogonal axes, $x$-axis $a_x$ and $z$-axis $a_z$, and estimate them separately.
The network learns confidence values to deal with the problem that the regressed two axes are not orthogonal:
\begin{align}
\begin{split}
\label{eq:rot}
    \left[\hat{a}_i, c_i\right] &= \mathcal{F}_i\left(\mathcal{P}_{concat}\right), \  i\in \left\{x, z\right\} \\
    \theta_z &= \frac{c_x}{c_x + c_z}\left(\theta - \frac{\pi}{2}\right)\\
    \theta_x &= \frac{c_z}{c_x + c_z}\left(\theta - \frac{\pi}{2}\right)
\end{split}
\end{align}
where $c_x, c_z$ denotes the confidence for the learned axes. $\theta$ represents the angle between $a_x$ and $a_z$.
$\theta_x, \theta_z$ are obtained by solving an optimization problem and then used to rotate $a_x$ and $a_z$ within their common plane.
For the training loss, we use $L_1$ plus cosine similarity loss $\mathcal{L}_{r_x}, \mathcal{L}_{r_z}$ for each individual axis. Angular loss, $\mathcal{L}_a$, is also used to constrain a perpendicular relationship between the two axes and confidence loss $\mathcal{L}_{con_x},\mathcal{L}_{con_z}$ to learn axis confidence:
\begin{align}
\begin{split}
\label{eq:rot_loss}
    \mathcal{L}_{r_i} &= \left\lvert\hat{a}_i - a^*_i\right\rvert + 1 - \left\langle\hat{a}_i, a^*_i\right\rangle, \  i\in \left\{x, z\right\} \\
    \mathcal{L}_a &= \left\langle\hat{a}_x, \hat{a}_z\right\rangle\\
    \mathcal{L}_{con_i} &= \left\lvert c_i - \text{exp}\left(\alpha \left\lVert \hat{a}_i - a^*_i \right\rVert_2\right)\right\rvert, \  i\in \left\{x, z\right\} 
\end{split}
\end{align}
where $\alpha$ is a constant to scale the distance. Thus the overall loss for the second stage is:
\begin{align}
\begin{split}
\label{eq:all_loss}
    \mathcal{L} = &\lambda_s\mathcal{L}_s + \lambda_t\mathcal{L}_t + \lambda_{r_x}\mathcal{L}_{r_x} + \lambda_{r_z}\mathcal{L}_{r_z} + \\
    &\lambda_{r_a}\mathcal{L}_{a} + \lambda_{con_x}\mathcal{L}_{con_x} + \lambda_{con_z}\mathcal{L}_{con_z}
\end{split}
\end{align}

where $\lambda_{r_x}, \lambda_{r_z}, \lambda_{r_a}, \lambda_{t}, \lambda_{s}, \lambda_{con_x}, \lambda_{con_z}$ are weights for each loss.


\noindent\textbf{Scale Residual Estimation}  
Similar to the translation decoder, we define the scale prior $s_{prior}$ as the average of scales of all object 3D CAD models within each category. Then the scale of a given instance is calculated as follows:
\begin{align}
\begin{split}
\label{eq:scale}
    &\hat{s} = s_{prior} + \mathcal{F}_{s}\left(\mathcal{P}_{concat}\right) \\
\end{split}
\end{align}

 The loss function is defined as the $L_1$ loss between the ground truth scale and estimated scale:

\begin{equation}
\label{eq:scale_loss}
    \mathcal{L}_s = \left\lvert\hat{s} - s^*\right\rvert
\end{equation}
\section{Experiments}
\label{sec:result}

\begin{figure}[h]
    \centering
    \includegraphics[width=0.95\columnwidth]{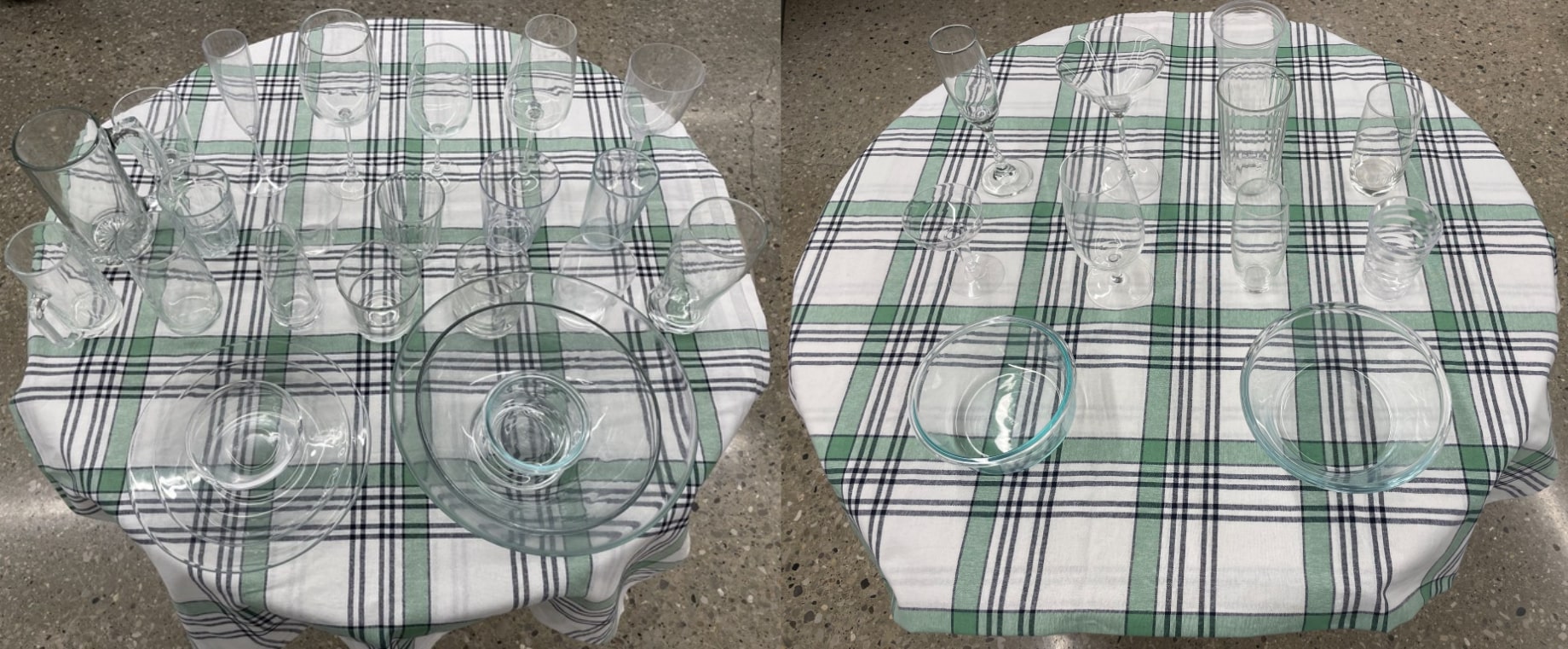}
    \caption{Transparent objects used in experiments. The image on the left show the 25 objects in the training set. The image on the right shows 8 novel objects in the test set for vision evaluation and robotic experiments.}
    \label{fig:objects}
\end{figure}

\noindent\textbf{Dataset} We evaluated TransNet and baseline models on the Clearpose Dataset \cite{chen2022clearpose} for depth completion, surface normal estimation, and categorical transparent object pose estimation tasks.
As shown in Figure~\ref{fig:objects}, 33 object instances from 3 ClearPose categories (\textit{bowl, water\_cup,} and \textit{wine\_cup}) were selected for evaluating the relevant models on each task. 25 of the objects are used in the training set, totaling 190K RGB-D images. Hence, 8 objects not seen during training are used for testing, with 60K test images available. In practice, we train the models using the full 190K image training set and uniformly sample images of the test object to form a representative 5K image test set. For the robot manipulation experiments, we used 1 \textit{bowl} object and 1 \textit{wine\_cup} object from the test set along with another 1 new \textit{wine\_cup} and 2 new \textit{water\_cups} purchased from the market that fit the robot parallel gripper.

\noindent\textbf{Implementation Details} Our model was trained in several stages. For experiments included in the ablation study and baseline comparison, ground truth instance segmentations are used to generate input patches, while for the robot experiments, a Mask R-CNN~\cite{he2017mask} model fine-tuned on the Clearpose dataset is used for generating input patches. 
The image patches were generated from object bounding boxes and re-scaled to a fixed shape of $256\times256$ pixels. 
Cross-task consistency training is performed using the AdamW optimizer~\cite{loshchilov2017decoupled} with learning rate of $1e^{-3}$. The pretrained surface normal estimation model provided by Zamir~\textit{et al.}~\cite{zamir2020robust} is used for initial network parameters.
Cross-task training hyperparameters follow the perceptual loss strategy described in~\cite{zamir2020robust} until convergence using a batch size of 8.
For TransCG, the AdamW optimizer~\cite{loshchilov2017decoupled} is used for training with $\lambda_{smooth} = 0.001$ and the overall learning rate is $0.001$ to train the model until converge. 
For U-Net, we used the Adam optimizer~\cite{kingma2014adam} with a learning rate of $1e^{-4}$ to train the model until convergence.
For both surface normal estimation and depth completion, the batch size was set to 24.
For the second stage, the training hyperparameters of Pointformer and pose and scale estimation were selected following~\cite{zou20216d, di2022gpv}.
The learning rate for all loss terms were kept the same during training, $\left\{\lambda_{r_x}, \lambda_{r_z}, \lambda_{r_a}, \lambda_{t}, \lambda_{s}, \lambda_{con_x}, \lambda_{con_z}\right\} = \left\{8, 8, 4, 8, 8, 1, 1\right\} e^{-4}$.
The Ranger optimizer \cite{liu2019variance,yong2020gradient,zhang2019lookahead} was used with a linear warm-up for the first 1000 iterations, then a cosine annealing method at the 0.72 anneal point was used.
All the experiments for pose estimation were trained on a 16G RTX3080 GPU until loss convergence.


\noindent\textbf{Evaluation metrics} For category-level pose estimation, this study follows \cite{di2022gpv,chen2021fs} in using 3D intersection over union (IoU) between the ground truth and estimated 3D bounding box 
at 25\%, 50\% and 75\% thresholds. Additionally, $5^{\circ}5cm$, $10^{\circ}5cm$, and $10^{\circ}10cm$ are used as metrics. The reported measurements for each of these three metrics represent the percentage of a model's pose estimates with error less than the specified metric's degree and distance thresholds. 

For depth completion evaluation, the root of mean squared error (RMSE), absolute relative error (REL) and mean absolute error (MAE) are used as metrics along with $\delta_{1.05}$, $\delta_{1.10}$, $\delta_{1.25}$, where $\delta_n$ is calculated as:

\begin{equation}
    \delta_n = \frac{1}{N_p}\sum_{p}\textbf{I}\left(\text{max}\left(\frac{\hat{\mathcal{D}}_p}{\mathcal{D}^*_p}, \frac{\mathcal{D}^*_p}{\hat{\mathcal{D}}_p}\right) < n\right)
\end{equation}

\noindent where $\textbf{I}(\boldsymbol{\cdot})$ represents the indicator function. $\hat{\mathcal{D}_p}$ and $\mathcal{D}^*_p$ represent estimated and ground truth depth for each pixel $p$.

For surface normal estimation, RMSE and MAE errors of normalized vectors are used as metrics with thresholds of $11.25^{\circ}$, $22.5^{\circ}$, and $30^{\circ}$. Here, measurements reported for the $11.25^{\circ}$ metric represent the percentage of estimates with an angular distance less than $11.25^{\circ}$ from the ground truth surface normals. The mean angular error in degrees (MEAN) is also reported.

\subsection{Comparison with Baseline}
\label{exp:baseline}
\setlength{\tabcolsep}{1.4pt}

\begin{figure}[h]
\vspace{0.2cm}
    \centering
    \includegraphics[width=0.95\columnwidth]{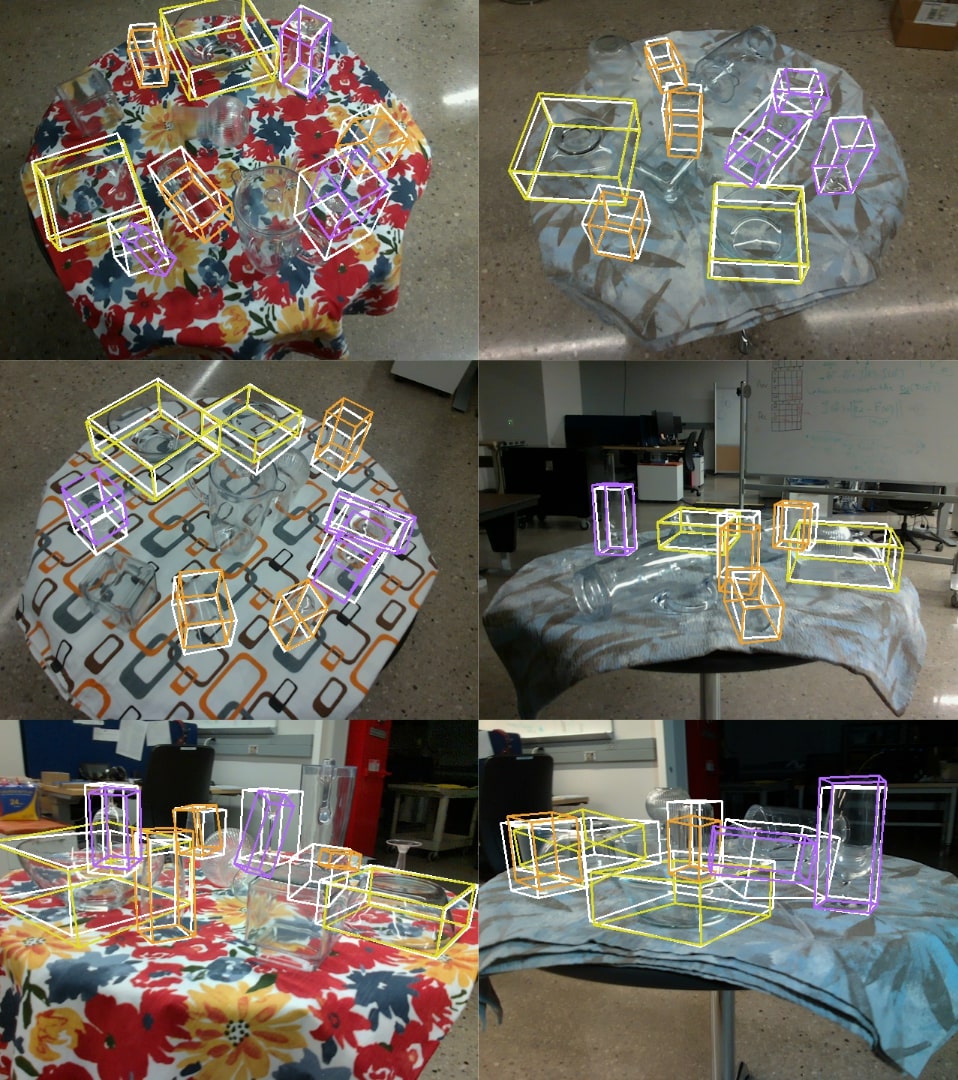}
    \caption{Qualitative results of category-level pose estimates from TransNet. White bounding boxes depict ground truth pose annotations, while colored boxes depict TransNet's estimates. Different colors represent different categories. 
     The top two rows show accurate estimates, while the bottom row highlights inaccurate estimates when TransNet is confronted with heavyily cluttered scenes.
     Objects with no bounding box correspond to categories not used during training and evaluation.
     {For axial symmetric objects, we use the ground truth x-axis to produce a 3D bounding box.}
     }
    \label{fig:visual}
\vspace{-0.2cm}
\end{figure}

We chose one state-of-the-art categorical opaque object pose estimation model (GPV-Pose \cite{di2022gpv}) as a baseline, which was trained with estimated depth from TransCG \cite{fang2022transcg} for a fair comparison. 
From Table \ref{table:baseline}, TransNet outperformed the baseline in most of the metrics on the Clearpose dataset. The $3\text{D}_{25}$ metric is relatively easy to achieve strong performance, so there is no substantial difference between the two models' corresponding $3\text{D}_{25}$ performance. 
When compared to the baseline on the remaining metrics, TransNet achieved around a 3 $\times$ improvement on $3\text{D}_{75}$, 3.5 $\times$ on $10^{\circ}5\text{cm}, 10^{\circ}10\text{cm}$ and 8 $\times$ on $5^{\circ}5\text{cm}$. Qualitative results are shown in Figure \ref{fig:visual} for TransNet. 

\setlength{\tabcolsep}{1pt}
\begin{table}[h]
\begin{center}
\caption{Comparison with the baseline on the Clearpose Dataset.}
\resizebox{\columnwidth}{!}{
\begin{tabular}{c|ccccccc}
\hline
\noalign{\smallskip}
Method & $\text{3D}_{25}\uparrow$ & $\text{3D}_{50}\uparrow$ & $\text{3D}_{75}\uparrow$ & $5^{\circ}5\text{cm}\uparrow$ & $10^{\circ}5\text{cm}\uparrow$ & $10^{\circ}10\text{cm}\uparrow$\\
\noalign{\smallskip} 
\hline
GPV-Pose & 95.4 & 65.1 & 13.2 & 2.7 & 12.5 & 15.5 \\
\hline
TransNet & \textbf{97.3} & \textbf{80.3} & \textbf{39.9} & \textbf{22.7} & \textbf{45.4} & \textbf{50.6}   \\
\hline
\end{tabular}
}
\label{table:baseline}
\end{center}
\vspace{-0.7cm}
\end{table}

\subsection{Ablation Study}

\label{exp:ablation}

In Table \ref{table:ablation}, we compared the performance of TransNet to that of modified versions of the network architecture. The results of this study informed Trial 6 as the final architecture of TransNet and the one that was used in robot experiments. Results from the ablation study may also be used to guide future work in the area of transparent object pose estimation. 

\begin{table*}
\vspace{0.2cm}
\begin{center}
\caption{Ablation study for TransNet}
\begin{tabular}{c|cc|cccc|c|c|ccccccc}
\hline
\multirow{2}{*}{Trial} &  \multicolumn{2}{c|}{Embedding} & \multicolumn{4}{c|}{Generalized Point Cloud} & \multirow{2}{*}{Consistency} & \multirow{2}{*}{Separate Category} & \multirow{2}{*}{$3D_{25}\uparrow$} & \multirow{2}{*}{$3D_{50}\uparrow$} & \multirow{2}{*}{$3D_{75}\uparrow$} & \multirow{2}{*}{$5^{\circ}5cm\uparrow$} & \multirow{2}{*}{$10^{\circ}5cm\uparrow$} & \multirow{2}{*}{$10^{\circ}10cm$$\uparrow$} \\ \cline{2-7}
& \multicolumn{1}{c}{PF}  & \multicolumn{1}{c|}{3D-GCN} & \multicolumn{1}{c}{RGB} & \multicolumn{1}{c}{Depth} & \multicolumn{1}{c}{Normal} & \multicolumn{1}{c|}{Ray-direction}  & & & & & & & & \\ \hline

\multicolumn{1}{c|}{1}&  & \checkmark & \checkmark & \checkmark &  & \checkmark & & & 98.3 & 72.7 & 17.6  & 6.6 & 22.5 & 28.0 \\ 

\multicolumn{1}{c|}{2} & \checkmark &   & \checkmark & \checkmark &  & \checkmark & &  & 97.9 & 76.9 & 28.4  & 9.4 & 8.0 & 29.6 \\ 

\multicolumn{1}{c|}{3}& \checkmark &  & \checkmark & \checkmark & \checkmark & & & &  95.4 & 61.6 & 11.6  & 3.6 & 15.3 & 31.3 \\ 

\multicolumn{1}{c|}{4}& \checkmark &   & \checkmark & \checkmark & \checkmark & \checkmark  & & & \textbf{98.4} & 81.6 & 34.4  & 13.0 & 38.6 & 45.1 \\ 

\multicolumn{1}{c|}{5}& \checkmark &  & \checkmark & \checkmark & \checkmark & \checkmark  & \checkmark & & 98.1 & \textbf{81.8} & 39.2  & 15.8 & 41.4 & 46.1 \\ 

\hline

\multicolumn{1}{c|}{6}& \checkmark &  & \checkmark & \checkmark & \checkmark & \checkmark  & \checkmark & \checkmark & 97.3 & 80.3 & \textbf{39.9} & \textbf{22.7} & \textbf{45.4} & \textbf{50.6} \\ 

\hline
\end{tabular}
\label{table:ablation}
\end{center}
\vspace{-0.5cm}
\end{table*}

\noindent\textbf{Embedding Method}
\label{exp:embedding}
Between trial 1 and trial 2, we compared the effect of the embedding method, 3D-GCN \cite{Lin_2020_3dgcn} and Pointformer \cite{zou20216d}, on TransNet accuracy. The use of Pointformer resulted in higher accuracy for TransNet on most metrics compared to when 3D-GCN was used. More details are shown in Table \ref{table:emb_ab}. Crucially, when the ground truth depth was changed to estimated depth (modeling the change from opaque to transparent setting), Pointformer retained substantially more accuracy than 3D-GCN. A possible explanation for this result is based on how 3D-GCN propagates information between nearest neighbors. While this is an efficient method for propagating information locally, it suffered when given the noisy completed depth as input. Pointformer, on the other hand, shares information across the whole point cloud, possibly contributing to increased robustness when input data is noisy. Therefore, given depth with large uncertainty, the transformer-based embedding method might be more powerful than embedding methods using nearest neighbors. 

\setlength{\tabcolsep}{1pt}
\begin{table}
\begin{center}
\caption{Comparison between different embedding methods}
\resizebox{\columnwidth}{!}{
\begin{tabular}{c|c|ccccccc}
\hline
Depth type & Embedding& $\text{3D}_{25}$$\uparrow$ & $\text{3D}_{50}$$\uparrow$ & $\text{3D}_{75}$$\uparrow$ & $5^{\circ}5\text{cm}$$\uparrow$ & $10^{\circ}5\text{cm}$$\uparrow$ & $10^{\circ}10\text{cm}$$\uparrow$\\
\hline
\multirow{2}{*}{Ground truth} 
& 3D-GCN & 99.8 & 96.6 & 55.4 & 70.9 & 86.0 & 90.3 \\
& Pointformer & 99.7 & 96.6 & 74.5 & 61.9 & 83.7 & 86.5 \\
\hline
\multirow{2}{*}{Estimation} 
& 3D-GCN & 98.3  & 72.7 & 17.6 & 6.6 & 22.5 & 28.0 \\
& Pointformer &97.9 & 76.9 & 28.4 & 9.4 & 29.6 & 35.3 \\
\hline
\end{tabular}
}
\label{table:emb_ab}
\end{center}
\vspace{-0.3cm}
\end{table}

\noindent\textbf{Modalities for Generalized Point Cloud}
\label{exp:generalized point cloud}
The comparison of trials 2—4 was used to identify the optimal combination of feature inputs for the generalized point cloud in TransNet. For trials 2 and 4, we compared the effect of adding the estimated surface normal to the generalized point cloud. 
Results from this comparison imply that surface normals serve as a good complement when depth estimation is not accurate, as in the transparent setting.
For trials 3 and 4, we compared the potential benefit of including ray-direction. 
Notably, the ray-direction input contains both camera intrinsic and pixel positional information, which we hypothesize is critical for the network to be able to model accurate 3D spatial relationships.
All the metrics showed that including both surface normal and ray-direction results in improved pose estimation accuracy.

\setlength{\tabcolsep}{1pt}
\begin{table}
\begin{center}
\caption{Accuracy for depth-normal pair estimation}
\begin{tabular}{c|cccccc}
\hline
Depth Metrics &RMSE$\downarrow$ & REL$\downarrow$ & MAE$\downarrow$ & $\delta_{1.05}$$\uparrow$ & $\delta_{1.10}$$\uparrow$ & $\delta_{1.25}$$\uparrow$\\
\hline
w/o consistency & \textbf{0.056} & \textbf{0.044} & \textbf{0.041} & 68.61 & 89.35 & \textbf{98.89}\\
w consistency  & 0.057 & \textbf{0.044} & \textbf{0.041} & \textbf{70.49} & \textbf{89.47} & 98.57 \\
\hline\hline
Surface Normal Metrics & RMSE$\downarrow$ & MAE$\downarrow$ & MEAN$\downarrow$ & $11.25^{\circ}$$\uparrow$ & $22.5^{\circ}$$\uparrow$ & $30^{\circ}$$\uparrow$\\
\hline
w/o consistency  & 0.19 & 0.13 & 11.43 & 56.75 & 88.45 & \textbf{96.64} \\
w consistency  & \textbf{0.12} & \textbf{0.08} & \textbf{8.96} & \textbf{73.62} & \textbf{92.56} & 96.58 \\
\hline
\end{tabular}
\label{table:d-s}
\end{center}
\vspace{-0.7cm}
\end{table}

\noindent\textbf{Consistent Learning}
For trials 4 and 5 in Table \ref{table:ablation}, TransNet is trained without and with cross-task consistency, respectively. For trial 4, both surface normal estimation and depth completion networks in TransNet's first stage are trained separately and without the $\mathcal{L}_{con}$ loss. Results from this comparison show that adding cross-task consistency training improved TransNet performance on high accuracy metrics: $5^{\circ}5\text{cm}$, $3\text{D}_{75}$. From Table \ref{table:d-s}, adding a cross-task consistency loss resulted in improved accuracy for depth-normal pair estimation, particularly for the surface normal estimation network. Therefore, cross-task consistent training improves the performance of category-level transparent object pose estimation and is included in TransNet. 

\noindent\textbf{Separate Category}
{As shown in Table \ref{table:ablation} when comparing results in trial 5 and trial 6, it was found that TransNet's accuracy improves when a single instantiation of the model is trained per category as opposed to being trained on multiple categories. Notably, this result is most apparent for the high accuracy metric $5^{\circ}5\text{cm}$}

\noindent\textbf{Why TransNet Outperforms GPV-Pose}
{ Comparing GPV-Pose with the trial 1 model leads to an observation that both models use 3D-GCN and a similar network architecture. The only difference is that instead of using the generalized point cloud as input, GPV-Pose directly uses a point cloud (the multiplication of depth and ray direction) as input. Yet, trial 1 outperforms GPV-Pose on pose estimation tasks. We hypothesize that the camera intrinsic and pixel positional information contained in ray-direction degrades in accuracy after being multiplied directly with noisy depth data to form the point cloud input to GPV-Pose. In contrast, TransNet retains ray-direction and depth information as distinct feature channels in the generalized point cloud with learned transformations to combine each feature modality. These learned transformations potentially allow TransNet to be more robust in the face of noisy depth data.}

\begin{figure}[h]
    \centering
    \includegraphics[width=\columnwidth]{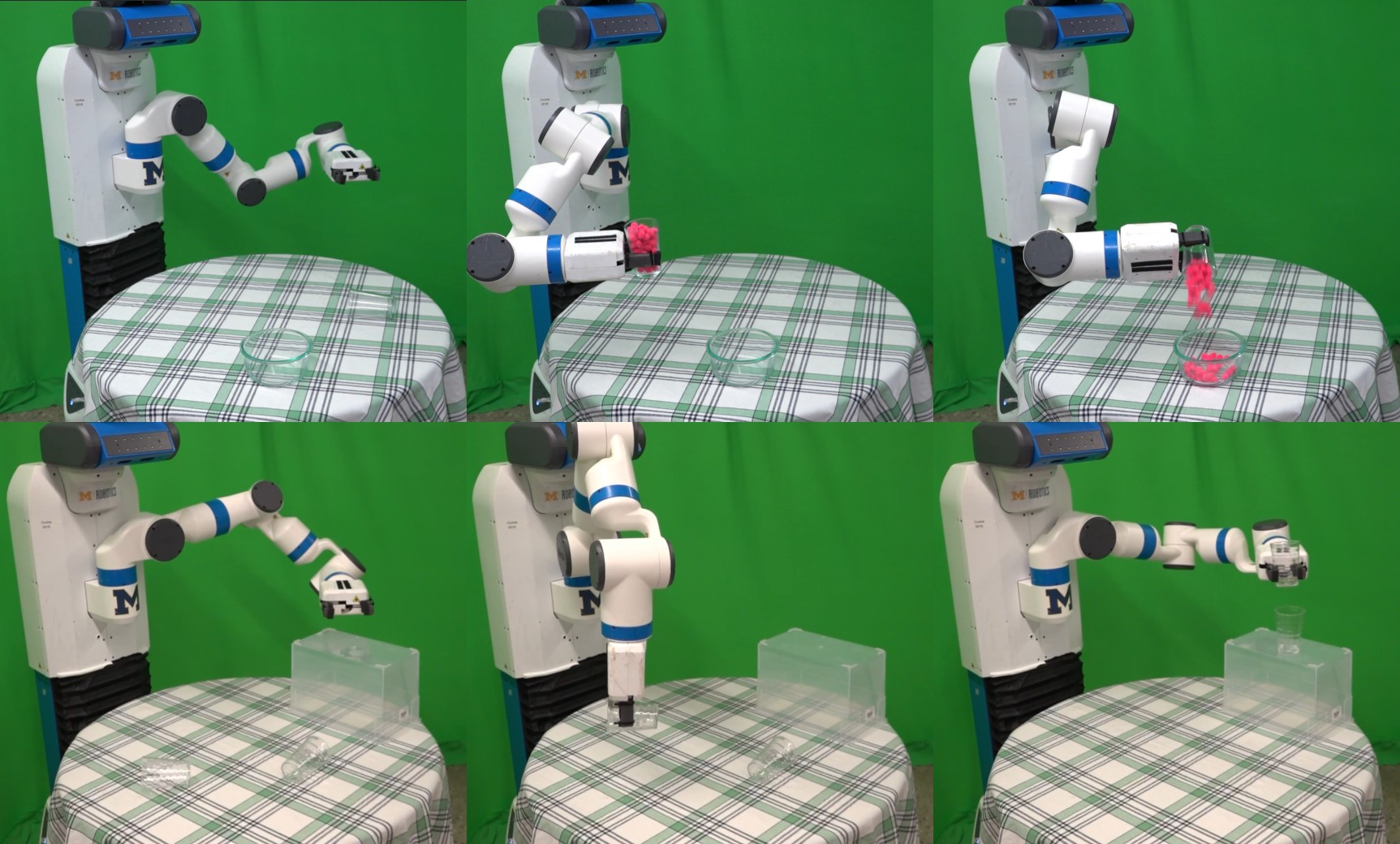}
    \caption{Robotic experiments. The top row is the demo of pouring using a water cup and a bowl. The bottom row is the demo of pick-and-place using two water cups. The left column is the initial state, the middle column is picking, and the right column is pouring or placing.}
    \label{fig:robotgrasping}
    \vspace{-0.5cm}
\end{figure}  

\subsection{Robot Experiment}

We used a Fetch robot for manipulation and an Intel RealSense L515 camera to take RGB-D images as used in the ClearPose dataset. The robot-camera extrinsic calibration is done using AprilTags~\cite{krogius2019iros}. Inspired from \cite{chen2020manipulation}, we designed two robot manipulation tasks: pouring and pick-and-place to evaluate the efficacy of TransNet to support robot manipulation of transparent objects using 3 \textit{water\_cups}, 2 \textit{wine\_cups}, and 1 \textit{bowl} not appearing in the training set. 
{For grasping the object, we calculate the approaching direction as orthogonal to the object's symmetrical axis and the grasp location as the object's center point.}

\noindent\textbf{Pick-and-place} 
This task is composed of a pick action and a place action. For the pick action, the robot is to grasp a cup (either \textit{water\_cup} or \textit{wine\_cup}) on the table-top with its parallel gripper based on its pose estimate from TransNet. For the place action, the robot is to place the cup upright in a pre-defined location. The target poses for the robot's gripper are solved by aligning the object's estimated bottom surface parallel to the target location surface with a 2 cm offset.

\noindent\textbf{Pouring} 
This task is composed of a pick action and a pouring action. The pick action follows the same protocol as \textit{pick-and-place}. For the pouring action, the robot moved the cup above the bowl based on the bowl's estimated location while also keeping the cup's $z$ axis upright to prevent spilling. Then the content in the cup is poured into the bowl by tilting the cup's $z$ axis.


\noindent\textbf{Result}
The poses and scales used by the robot are output directly from TransNet without additional post-processing. Quantitative results measuring the robot's success rate are included in Table~\ref{table:robotgrasp}. 
TransNet enabled the robot to perform pour and pick-place actions with promising success rates. Visualization examples are included in Figure \ref{fig:robotgrasping}.


\begin{table}
\centering
\caption{Robot experiment success rate}
\resizebox{\columnwidth}{!}{
\begin{tabular}{c|c|cc|cc}
\hline
Task & Objects   & \#Pick & Success Rate & \#Place(Pour) & Success Rate \\ \hline
\multirow{2}{*}{Pick \& Place} & water\_cup         & 16/20  & 80\%         & 12/16         & 75\%         \\
                               & wine\_cup          & 13/20  & 65\%         & 12/13         & 92.3\%       \\ \hline
\multirow{2}{*}{Pouring}       & water\_cup \& bowl & 15/20  & 75\%         & 10/15         & 66.7\%       \\
                               & wine\_cup \& bowl  & 14/20  & 70\%         & 11/14         & 78.6\%       \\ \hline
\end{tabular}}
\label{table:robotgrasp}
\vspace{-0.5cm}
\end{table}


\section{Conclusions}
\label{sec:conclusion}

In this paper, we introduced \textit{TransNet}, a two-stage pipeline for category-level transparent object pose estimation that outperformed the
baselines. Ablation studies were performed to understand the benefits of using multi-modal input, feature embedding modules, and cross-task consistency. The efficacy of TransNet was demonstrated by enabling a robot system to perform transparent object manipulation tasks with high success rates. 
In the future, we could further improve the pose estimation accuracy as there is still a large gap between TransNet and state-of-the-art opaque object datasets. One direction is to consider objects' material information, as we observed a performance gap between perceiving glass and plastic objects.


\bibliographystyle{IEEEtran}
\bibliography{egbib}

\end{document}